%% file: main.tex
\documentclass{article}
\usepackage{geometry}
\usepackage{amsmath}
\usepackage{amssymb}  
\usepackage[export]{adjustbox}
\usepackage{microtype}
\usepackage{graphicx}
\usepackage{subfig}
\usepackage{booktabs}
\usepackage{xr}
\usepackage{authblk}
\usepackage{natbib}
\usepackage{algorithmic}
\usepackage{algorithm}

\usepackage{hyperref}
\hypersetup{
    colorlinks,%
     citecolor=blue,%
     filecolor=blue,%
     linkcolor=red,%
     urlcolor=blue
}

\newtheorem{theorem}{Theorem}
\newtheorem{corollary}{Corollary}
\newtheorem{lemma}{Lemma}

\renewcommand{\algorithmiccomment}[1]{#1}

\def\cA{\mathcal{A}}
\def\cH{\mathcal{H}}
\def\cN{\mathcal{N}}
\def\bE{\mathbb{E}}
\def\bP{\mathbb{P}}
\def\calH{\mathcal{H}}
\def\indc#1{\mathbf{1}\{#1\}}
\def\argmax#1{{\arg\!\max}_{#1}}

\newcommand{\regret}{\textnormal{Regret}}
\makeatletter
\newcommand*{\addFileDependency}[1]{
  \typeout{(#1)}
  \@addtofilelist{#1}
  \IfFileExists{#1}{}{\typeout{No file #1.}}
}
\makeatother

\newcommand*{\myexternaldocument}[1]{%
    \externaldocument{#1}%
    \addFileDependency{#1.tex}%
    \addFileDependency{#1.aux}%
}
\myexternaldocument{supplement}

\title{Online Multi-Armed Bandits with Adaptive Inference}

\author[1]{Maria Dimakopoulou}
\author[2]{Zhimei Ren}
\author[3]{Zhengyuan Zhou}
\affil[1]{Netflix}
\affil[2]{Stanford University}
\affil[3]{NYU Stern School of Business}
\date{}

\begin{document}

\maketitle

\begin{abstract}
During online decision making in Multi-Armed Bandits (MAB), one needs to conduct inference on the true mean reward of each arm based on data collected so far at each step. However, since the arms are adaptively selected--thereby yielding non-iid data--conducting inference accurately is not straightforward. In particular, sample averaging, which is used in the family of UCB and Thompson sampling (TS) algorithms, does not provide a good choice as it suffers from bias and a lack of good statistical properties (e.g.  asymptotic normality). Our thesis in this paper is that more sophisticated inference schemes that take into account the \textit{adaptive} nature of the sequentially collected data can unlock further performance gains, even though both UCB and TS type algorithms are optimal \textit{in the worst case}. In particular, we propose a variant of TS-style algorithms--which we call doubly adaptive TS--that leverages recent advances in causal inference and \textit{adaptively} reweights the terms of a doubly robust estimator on the true mean reward of each arm. Through 20 synthetic domain experiments and a semi-synthetic experiment based on data from an A/B test of a web service, we demonstrate that using an adaptive inferential scheme (while still retaining the exploration efficacy of TS) provides clear benefits in online decision making: the proposed DATS algorithm has superior empirical performance to existing baselines (UCB and TS) in terms of regret and sample complexity in identifying the best arm. In addition, we also provide a finite-time regret bound of doubly adaptive TS that matches (up to log factors) those of UCB and TS algorithms, thereby establishing that its improved practical benefits do \textit{not} come at the expense of worst-case suboptimality. 
\end{abstract}

\section{Introduction}

Stochastic Multi-Armed Bandits (MAB)~\citep{robbins1952some, berry1985bandit, bubeck2012regret, sutton2018reinforcement,lattimore2020bandit} is the simplest and most well-established model for sequential decision making, where a decision maker needs to adaptively select an arm at each time from a set of arms with unknown (but fixed) mean rewards, in the hope of maximizing the total accumulated expected returns over a certain time horizon. The existing literature has studied this problem extensively, mainly focusing on developing algorithms that deal with the exploration-exploitation trade-off, yielding at least two broad classes of algorithms that provide optimal (sometimes up to log factors) regret guarantees \textit{in the worst case}.

The first is upper confidence bound (UCB) based algorithms~\citep{lai1985asymptotically, agrawal_1995, auer2002finite,auer2002using,garivier2011upper, garivier2011kl,carpentier2011upper}.
Reflecting ``optimism in face of uncertainty", UCB algorithms compute confidence bounds of the estimated mean, construct the index for each arm by adding the confidence bound to the mean (as the best statistically plausible mean reward) and select the arm with the highest index. The finite-time regret bounds for these algorithms--$\Theta(\log T)$ gap-dependent bounds and $\Theta(\sqrt{T})$ gap-independent bounds--are optimal in the worst case.

On the other hand, UCB algorithms are known to be sensitive to hyper-parameter tuning, and are often hard to tune for stellar performance. This stands in contrast to Thompson sampling (TS) based algorithms~\citep{thompson1933likelihood, kaufmann2012thompson, ghavamzadeh2015bayesian, russo2017tutorial}, an algorithm that does not require much tuning and that achieves exploration through ``probabilistic optimism in face of uncertainty". Further, as a result of this more nuanced exploration scheme, TS has been widely recognized to outperform UCB in empirical applications~\citep{scott2010modern,graepel2010web, may2011simulation, chapelle2011empirical}. However, although an old algorithm~\citep{thompson1933likelihood}, its finite-time worst-case regret bound was not known at the time. Consequently, driven by its empirical performance, its theoretical guarantee was raised as an open problem in COLT 2012~\citep{li2012open} and was subsequently settled by~\citep{agrawal2012analysis,agrawal2013further}, which yield the same minimax optimal regret guarantees (up to log factors) as in UCB. These developments---both the UCB/TS algorithms and the theoretical guarantees---have subsequently been successfully applied to contextual bandits\footnote{And to reinforcement learning as well, but contextual bandits and RL are not the focus of this paper.}~\citep{LCLS2010,FCGS2010, chu2011contextual,JBNW2017, LLZ2017, AG2013a, AG2013b, RV2014, russo2016information,agrawal2017thompson}.

Despite this remarkably fruitful line of work, searching for better MAB algorithms is far from over. For one thing, minimax optimality--optimality in the worst case--is often too conservative a metric, and does not serve as a useful indicator of practical performance for average problems\footnote{The fact that TS performs better in practice than UCB, even though the existing UCB bounds are tighter (e.g. by log factors) than those of TS, already attests to that. For instance, \cite{audibert2009minimax} showed that a variant of UCB has regret $O(\sqrt{KT})$ in MAB while the best known regret bound for TS is $O(\sqrt{KT\log T})$~\citep{agrawal2017near}. And for contextual bandits, TS is often at least a factor of  $d$ (context dimension) worse.}. In particular, an important weakness in UCB/TS algorithms is that the true mean reward of each arm is estimated using a simple sample average, which would work if the data \textit{were} iid generated. However, the data is adaptively collected and consequently, as pointed out in~\citep{xu2013estimation,luedtke2016statistical,bowden2017unbiased,nie2018adaptively,neel2018mitigating, hadad2019confidence,shin2019bias} directly using the sample average to estimate the true mean rewards creates large bias. Intuitively, this is because arms which appear to have worse reward performance than they actually do due to randomness are sampled less often and the downward bias is not corrected.

As such, the inferential quality of an estimator--when treating the underlying data \textit{as if} they are generated iid--will be poor\footnote{For instance, sample average is not even asymptotically normal under adaptively collected data.}, thereby constraining the overall online decision making performance.

Our goal in this paper is to leverage recent advances in the causal inference literature~\citep{luedtke2016statistical, hadad2019confidence} and obtain practically better online performance by harnessing the strengths of both the adaptive inference estimators\footnote{Such as the estimator proposed by \citep{luedtke2016statistical} and later generalized by \citep{hadad2019confidence}. \citep{luedtke2016statistical} and \citep{hadad2019confidence} studied the offline estimation problem from adaptively collected data and the focus of these papers was to estimate
the true mean rewards accurately by providing a bias-corrected estimation based on the doubly-robust estimator \citep{imbens2015causal}, but with variance-stabilizing weights that adapt to the decision history of the completed experiment.     } (originally designed for offline, post-experiment analyses of adaptive experiments) and the effective exploration-exploitation balance provided by TS, thereby designing a more effective \textit{online} sequential decision making algorithm.

\subsection{Our Contributions}
\label{sec:contribution}
Our contributions are threefold. First, we pinpoint the issues of the existing bandit algorithms (Section~\ref{subsec:motivation}) and design a new algorithm--which we call doubly-adaptive Thompson sampling (DATS)--by incorporating the adaptively-weighted doubly robust estimator in~\citep{luedtke2016statistical, hadad2019confidence} into the online decision making process and making the necessary modifications to the estimated variance of that estimator to render it suitable for the exploration/exploitation trade-off and ensure sufficient exploration. This algorithm mitigates the issues of poor inferential quality inherent in the sample average used in TS/UCB, while retaining the benefits of intelligent exploration (see Section~\ref{subsec:algorithm} for a more detailed discussion). Previously, IPW has been used in bandits\citep{agarwal2014taming}, motivated, analyzed and evaluated in \citep{dimakopoulou2017estimation,dimakopoulou2019balanced} and systematically benchmarked in \citep{bietti2018contextual}. However, IPW has very high variance and these works did not use variance stabilizing weights, which can yield poor performance--as we will see in section \ref{semisynthetic}. 
Second, we validate that the desired benefits in the design of DATS do translate into higher quality  decision making. In particular, through 20 synthetic domain experiments and one semi-synthetic experiment, we demonstrate that DATS is more effective than TS and UCB. This effectiveness is shown in two metrics: one is regret, which measures the cumulative performance compared to that of the optimal oracle; the other is sample complexity, which measures how many rounds are needed in order to identify the best arm. Under both metrics, DATS beats all existing bandit algorithms with a clear margin (see Section~\ref{sec:empirical} for a detailed presentation of results).
Finally, to complete the picture, we show that DATS achieves this superior practical performance without giving away the comparable worst-case regret guarantees. In particular, we establish that DATS has a finite-time regret of $O\big(K^2\sqrt{T\log{T}}\big)$, which is minimax optimal (up to log factors) in the horizon $T$, thus matching that of TS ($K$ is the number of arms)

Notably, all existing results on adaptive estimators in the inference literature are asymptotic in nature (typically in the style of central limit theorem bounds), whereas our bound is finite in nature and sheds light in the arena of online decision making, rather than offline inference, on which all prior work on adaptive estimators has focused--to our knowledge.  We point out that we do not attempt to be tight in $K$ in our bound, as in the applications we have in mind (e.g. clinical trials, web-service testing), including the one motivating our semi-synthetic experiment, the number of arms is typically a (small) constant. That said, our synthetic empirical results (Fig.~\ref{fig:synthetic}) on 20 domains (including for large $K$) demonstrate that DATS indeed has an optimal $O(\sqrt{K})$ dependence. We leave tighter analysis on $K$ for future work.

\section{Problem Formulation}
\label{problem}
In stochastic MABs, there is a finite set $\cA$ of arms $a \in \cA$ with $|\cA| = K$.
At every time $t$, the environment generates--in an iid manner--a reward $r_t(a)$ for every arm $a \in \cA$ with expectation $\bE[r_t(a)] = \mu_a$, where $\mu_a$ is the unknown true reward of arm $a$ which is modified by mean-zero noise to generate $r_t(a)$.
The optimal arm is the arm with the maximum true reward, which is denoted by $a^* := \arg\max_{a \in \cA} \mu_a$ and is also unknown.
When at time $t$ the decision maker chooses arm $a_t$, \textit{only} the reward of the chosen arm $r_t := r_t(a_t)$ is observed.

At every time $t$, the decision maker employs a policy $\pi_t$ that maps the history of arms and rewards observed up to that time, $\cH_{t-1} = (a_0, r_0, \dots, a_{t-1}, r_{t-1})$, to a probability distribution over the set of arms $\cA$ and chooses $a_t \sim \pi_t$. The probability with which arm $a$ is chosen at time $t$ (often referred to as propensity score of arm $a$ at time $t$) is $\pi_{t, a} = \bP(a_t = a | \cH_{t-1})$.
The goal of the decision maker is to make decisions adaptively and learn a sequence of policies $(\pi_1, \dots, \pi_T)$ over the duration of $T$ time periods, so that the expected cumulative regret with respect to the optimal arm selection strategy over $T$ time periods is minimized, where $\text{Regret}(T,\pi) := \sum_{t=1}^{T} \left(\mu_{a^*} - \mu_{a_t}\right)$.

\subsection{Baseline Algorithms: UCB and TS}
\label{baselines}
Upper confidence bound (UCB) is an algorithm that balances the exploration/exploitation trade-off by forming for each arm $a \in \cA$ at every time period $t$ an upper bound $U_{t, a}$ that represents the maximum statistically plausible value of the unknown true reward $\mu_a$ given the history $\cH_{t-1}$. 
Then, at time $t$, UCB chooses the arm with the highest bound, $a_t = \arg\max_{a \in \cA} U_{t, a}$. Hence, the policy $\pi_t$ of time $t$ assigns probability one to the arm with the highest bound (breaking ties deterministically) and probability zero to all other arms. 
An example is UCB-Normal for normally distributed
rewards \citep{auer2002finite}, where the upper confidence bound of arm $a$ at time $t$ is given by $U_{t, a} = \bar{r}_{t, a} + \beta \sqrt{ \hat{\sigma}^2_{t, a} \log(t-1)}$, where $\bar{r}_{t, a}$ is the sample average and $\hat{\sigma}^2_{t, a} = \frac{q_{t, a} - n_{t, a} \bar{r}_{t, a}}{n_{t, a}(n_{t, a} - 1)}$ is an estimate of the mean reward's variance ($q_{t, a}$ is the sum of squared rewards and $n_{t,a}$ is the number of pulls of arm $a$ up to time $t$) while $\beta$ is an algorithm parameter.
Thompson sampling (TS) is another algorithm that balances the exploration/exploitation trade-off by forming for each arm $a \in \cA$ at every time period $t$ a posterior distribution $\bP(\mu_a \in \cdot | \cH_{t-1})$, drawing a sample $\tilde{r}_{t, a}$ from it and choosing the arm with the highest sample, $a_t = \arg\max_{a \in \cA} \tilde{r}_{t, a}$. 
Hence, the policy $\pi_t$ of time $t$ is to choose each arm with the probability that it is optimal given the history of observations $\cH_{t-1}$, $\pi_t(a) = \bP(a =\arg\!\max_{a\in\cA}\tilde{r}_{t,a} | \cH_{t-1})$.
An example is the TS-Normal, in which it is assumed that the true reward $\mu_a$ of arm $a$ is drawn from $\cN(\hat{\mu}_{0, a}, \hat{\sigma}^2_{0, a})$, which plays the role of a prior distribution, and the realized reward of arm $a$ at time $t$, $r_t(a)$, is drawn from $\cN(\mu_a, \sigma^2)$, where $\mu_a$ is unknown and $\sigma$ is known. The posterior distribution of arm $a$ at time $t$ is also Normal with mean $\hat{\mu}_{t, a} = \bE[\mu_a | \cH_{t-1}] = (\bar{r}_{t, a} \hat{\sigma}^2_{t-1, a} +  \hat{\mu}_{t-1, a} \sigma^2 / n_{t, a}) / (\hat{\sigma}^2_{t-1, a} + \sigma^2 / n_{t, a})$ and variance $\hat{\sigma}^2_{t, a} = \bE[(\mu_a - \hat{\mu}_{t, a})^2  | \cH_{t-1}] = (\hat{\sigma}^2_{t-1, a}  \sigma^2 / n_{t, a}) / (\hat{\sigma}^2_{t-1, a} + \sigma^2 / n_{t, a})$ \citep{russo2017tutorial}.

The sample average $\bar{r}_{t, a}$ of arm's $a$ observed rewards up to time $t$ plays a prominent role in UCB and TS, since it is used in $U_{t,a}$ and $\bP(\mu_a \in \cdot | \cH_{t-1})$ respectively. 
However, the observations of a MAB algorithm are adaptively collected and, as a result, they are not independent and identically distributed, which makes these sample averages biased. This challenge is shared by both UCB and TS and motivates the MAB algorithm we propose in section \ref{algorithm}.

\section{Doubly-Adaptive Thompson Sampling}
\label{algorithm}

\subsection{Motivation}\label{subsec:motivation}
The issue with the sample average $\bar{r}_{t,a} = \frac{\sum_{s=1}^t \textbf{1}(a_s = a) r_s}{\sum_{s=1}^t \textbf{1}(a_s = a)}$ in the posterior $\bP(\mu_a \in \cdot | \cH_{t-1})$ or in the upper confidence bound $U_{t, a}$ of arm $a$ at time $t$ is that the sample average from adaptively collected data is neither unbiased nor asymptotically normal~\citep{xu2013estimation,luedtke2016statistical,bowden2017unbiased,nie2018adaptively,hadad2019confidence,shin2019bias}. 
One approach to correct the bias is to use inverse propensity score weighting (IPW), $\hat{Q}^{\text{IPW}}_{t, a} = \frac{1}{t}\sum_{s=1}^t \frac{\textbf{1}(a_s = a)}{\pi_{s, a}}  r_s$, which gives an unbiased estimate of arm's $a$ true mean reward $\mu_a$ if the propensity scores are accurate.  
However, inverse propensity score weighting comes at the expense of high variance---particularly when the propensity scores become small. 
A related approach is the doubly-robust estimator (DR),  $\hat{Q}^{\text{DR}}_{t, a} =  \frac{1}{t}\sum_{s=1}^t \left[ \bar{r}_{s-1,a} + \frac{\textbf{1}(a_s = a)}{\pi_{s, a}}\left(r_s -  \bar{r}_{s-1,a}\right)\right]$, which uses the sample average as the baseline and applies inverse propensity score weighting to a shifted reward using the sample average as a control variate. If the propensity scores are accurate, then DR is also unbiased and has improved variance compared to IPW, albeit still high. 

Also, both estimators fail to satisfy the variance convergence property \citep{hadad2019confidence} which is necessary for the central limit theorem \citep{hall2014martingale}.
Thus, IPW and DR--apart from high-variance--are also not asymptotically normal.

In order to stabilize the variance, \citep{luedtke2016statistical,hadad2019confidence,zhang2020inference} propose approaches of modifying the DR estimator by weighing the efficient score  $\hat{\Gamma}_{s,a} := \bar{r}_{s-1,a} + \frac{\textbf{1}(a_s = a)}{\pi_{s, a}}\left(r_s -  \bar{r}_{s-1,a}\right)$ of each data point $s \in [t] : =\{1, \dots, t\}$ by a non-uniform weight $w_{s, a}$ instead of weighing all data points uniformly with weight $1/t$. The resulting class of estimators is a weighted average of the efficient scores, which we refer to as adaptive doubly-robust (ADR), since these weights $w_{s, a}$ are \textit{adapted} to the history $\cH_{s-1}$. ADR takes the form $\hat{Q}^\text{ADR}_{t, a} = \frac{\sum_{s=1}^t w_{s, a} \hat{\Gamma}_{s,a}}{\sum_{s=1}^t w_{s, a}}$. Although the aforementioned works vary in terms of the weighting schemes they propose, their common intuition is that if the contribution of each term $\Gamma_{s,a}$ to the variance of the DR estimator is unequal, then using uniform weights is not as efficient (i.e., results in larger variance) as using weights inversely proportional to the standard deviation of each term $\Gamma_{s,a}$. Since the data collection at time $s$ adapts to the history $\cH_{s-1}$ via the propensity score $\pi_{s, a}$, so does the variance of $\Gamma_{s,a}$. As a result, the chosen weight $w_{s, a}$ is also \textit{adaptive} to the history $\cH_{s-1}$. \cite{luedtke2016statistical}, for instance, proposed weighing $\Gamma_{s,a}$ by $w_{s, a} = \frac{\sqrt{\pi_{s, a} / t}}{\sum_{s' = 1}^t \sqrt{\pi_{s', a} / t}}$. Subsequently, \cite{hadad2019confidence}, proposed a generalized mechanism for constructing weights $w_{s, a}$ such that the conditions of infinite sampling, variance convergence and bounded moments are satisfied, which are in turn necessary in order for the resulting ADR estimator to be unbiased, have low variance and an asymptotically normal distribution. 

\subsection{Algorithm}\label{subsec:algorithm}

\begin{algorithm}[t]
\caption{Doubly-Adaptive Thompson Sampling}
\label{alg:DATS}
\begin{algorithmic}
 
 \STATE \hspace{-10pt} \begin{tabular}{ l l }
 \textbf{Input:} & propensity threshold $\gamma \in (0, 1)$ (default 0.01)  
\end{tabular}

  \vspace{2pt}
  \FOR{$a \in \cA$}
  \STATE Pull arm $a$ and observe reward $r_{\text{initial}, a}$ and initialize $\bar{r}_{0, a} \leftarrow r_{\text{initial}, a}$, \enskip  $n_{0, a} \leftarrow 1$, $\pi_{1, a} \leftarrow 1 / K$.
  \ENDFOR
  \STATE $\cA_1 \leftarrow \cA$
  \vspace{2pt}
   \FOR{$t=1$ {\bfseries to} $T$}
   \vspace{2pt}
   \STATE Pull arm $a_t \sim \text{Multinomial}\left(\cA_t, (\pi_{t, a})_{a \in \cA_t}\right)$ and observe reward $r_t \leftarrow r_t(a_t)$.
    \vspace{4pt}
    \STATE \(\triangleright\) Update sample averages and counts.
    \STATE $\bar{r}_{t, a_t} \leftarrow \frac{n_{t-1, a_t} \bar{r}_{t-1, a_t} + r_t}{n_{t-1, a_t} + 1}$, \enskip $n_{t, a_t} \leftarrow n_{t, a_t} + 1$ and $\bar{r}_{t, a} \leftarrow \bar{r}_{t-1, a}$, \enskip $n_{t, a} \leftarrow n_{t-1, a}$ $\forall a \neq a_t$
  
  \vspace{4pt}
    \STATE \(\triangleright\) Update sampling distributions.
    \FOR{$a \in \cA_{t}$}
    \STATE 
    $\hat{\Gamma}_{s,a} \leftarrow \bar{r}_{s-1, a} + \indc{a_s = a}\frac{r_s - \bar{r}_{s-1,a}}{\pi_{s,a}}$, $\forall s \in [t]$,
    \STATE $\hat{\mu}_{t, a} \leftarrow \frac{\sum_{s=1}^t \sqrt{\pi_{s, a}}\hat{\Gamma}_{s,a}}{\sum_{s = 1}^{t} \sqrt{\pi_{s, a}}}$, \enskip
    $\hat{\sigma}^2_{t, a} \leftarrow \frac{\sum_{s=1}^t \pi_{s,a}\left[\hspace{-1pt}\left(\hat{\Gamma}_{s,a}- 
    \hat{\mu}_{t, a}\right)^2+ 1\hspace{-2pt}\right]}{\left(\sum_{s=1}^t \sqrt{\pi_{s, a}}\right)^2}$, $\forall s \in [t]$
    \ENDFOR
    
     \vspace{4pt}
    \STATE \(\triangleright\) Perform arm elimination.
   \vspace{2pt}
    \STATE $p_{t+1,a} \leftarrow \min_{a'\in\cA_t} \Phi\left(\frac{\hat{\mu}_{t, a} - \hat{\mu}_{t, a'}}{\sqrt{\hat{\sigma}^2_{t, a} + \hat{\sigma}^2_{t, a'}}}\right)$ and $\cA_{t+1} \leftarrow \cA_{t} - \{a \in \cA_{t}: p_{t+1, a} < 1/T \}$
     \vspace{4pt}
    \STATE \(\triangleright\) Compute propensity scores.
   \vspace{2pt}
    \STATE $\tilde{r}_{t+1,a} \sim\cN (\hat{\mu}_{t,a},\hat{\sigma}_{t,a}^2)$ and $\pi_{t+1, a} \leftarrow \mathbb{P}\Big(a = \argmax{a' \in \cA_{t+1}}~\tilde{r}_{t, a} \mid \cH_t\Big)$, $\forall a \in \cA_{t+1}$
    \STATE $\pi_{t+1, a} \leftarrow (1 - \gamma) \pi_{t+1, a}+ \gamma/|\cA_{t+1}|$, $\forall a \in \cA_{t+1}$
    \ENDFOR
\end{algorithmic}
\end{algorithm}
We propose \textit{doubly-adaptive Thompson sampling} (DATS), which uses as building block the ADR estimator  $\hat{Q}^\text{ADR}_{t, a} = \frac{\sum_{s=1}^t \sqrt{\pi_{s, a}} \hat{\Gamma}_{s,a}}{\sum_{s=1}^t \sqrt{\pi_{s, a}}}$ from \citep{luedtke2016statistical}.
Note that in online learning, the decision maker knows {\em exactly} the propensity scores in $\hat{Q}^\text{ADR}_{t, a}$. Additionally, in TS-based algorithms, these propensity scores are non-trivial and are bounded away from zero and one---at least during the initial stages of learning---unlike UCB, in which the propensity score is equal to one for the chosen arm and equal to zero for all other arms. With TS, at any time $t$ the propensity scores $\pi_{t, a}$ can be computed analytically or via Monte Carlo sampling based on the sampling distribution of each arm $a$ and then logged in order to be used for inference in subsequent time steps rather than the need to fit a propensity model in the UCB setting, which may not be accurate. For this important reason, our new MAB algorithm belongs to the TS class rather than the UCB class.

At each time step $t\in[T]$, where $T$ is the learning horizon, DATS forms a reward sampling distribution for each arm $a \in \cA$ based on the history of observations $\cH_{t-1}$ collected so far. 
The reward sampling distribution of arm $a$ at time $t$ is chosen to be normal $\cN(\hat{\mu}_{t,a}, \hat{ \sigma}^2_{t,a})$ with mean $\hat{\mu}_{t,a} := \hat{Q}^\text{ADR}_{t, a} =  \frac{\sum_{s=1}^t \sqrt{\pi_{s, a}}  \hat{\Gamma}_{s,a}}{\sum_{s=1}^t \sqrt{\pi_{s, a}}}$ and variance $\hat{ \sigma}^2_{t,a} := \frac{\sum^t_{s=1}\pi_{s,a}[(\hat{\Gamma}_{s,a} - \hat{\mu}_{t,a})^2+1]}{\left(\sum^t_{s=1} \sqrt{\pi_{s, a}} \right)^2}$. 
Note, that the variance in the sampling distribution of DATS has an auxiliary `+1' term compared to the variance in \citep{luedtke2016statistical,hadad2019confidence}.

This is essentially important for the online setting, as the auxiliary term guarantees a lower bound of the estimated variance, thereby ensuring sufficient exploration in the initial stages of learning (this is in contrast to the offline setting in~\citet{hadad2019confidence}). The derivation of $\hat{\sigma}_{t,a}$ used by DATS can be found in the Supplemental Material section \ref{sec:proof_good_event}.

At time $t$, the sampling distributions of each arm $a \in \cA$ are used to compute the probability of it being chosen, i.e., its propensity score. Given that TS is a probability matching heuristic, i.e., it plays an arm with the probability that it is optimal according to the reward sampling distribution, the propensity score of arm $a$ at time $t$ is equal to the probability with which a single sample $\tilde{r}_{t,a} \sim \cN(\hat{\mu}_{t,a}, \hat{ \sigma}^2_{t,a})$ from the sampling distribution of arm $a$ is greater than a single sample $\tilde{r}_{t,a'} \sim \cN(\hat{\mu}_{t,a'}, \hat{ \sigma}^2_{t,a'})$ from the sampling distribution of any other arm $a'$. Hence, $\pi_{s,a} = \bP\Big(\tilde{r}_{t,a} = \argmax{a'\in \cA_{t}} ~ \tilde{r}_{t,a'}\mid \calH_{t-1}\Big)$, where $\tilde{r}_{t,a} \sim\cN (\hat{\mu}_{t,a},\hat{\sigma}_{t,a}^2)$. These propensity scores can be computed with high accuracy via Monte Carlo simulation. In order to control the variance of their estimator, \citep{luedtke2016statistical} derived the asymptotic normality of their $\hat{Q}^\text{ADR}$ under the assumption that each arm has a non-negligible probability of being chosen, i.e., the propensity score of arm $a$ at every time $t$ is lower bounded---this however cannot be assumed in the online setting since otherwise the sub-optimal arms will be pulled for a non-trivial fraction of times and yield undesired regret. In order to deal with diminishing propensity scores in the online setting and derive a regret upper bound, DATS eliminates an arm $a$ from the set of eligible arms if there is another eligible arm $a'$ such that $\bP(\tilde{r}_a > \tilde{r}_{a'}\mid\calH_{t-1})$ falls below $1 / T$---that is, arm $a$ is dominated by $a'$---while maintaining a level of uniform exploration $\gamma$ among the non-eliminated arms. $\gamma$ is a parameter of the algorithm (default $\gamma = 0.01$) and controls how small propensity scores get. The set of arms available to the decision maker are updated at each time step $t$ after the successive elimination of arms is denoted by $\cA_{t}$.

\section{Empirical Evaluation}
\label{sec:empirical}

We now present computational results that demonstrate the robustness of DATS in comparison to baselines, both in terms of regret and in terms of sample complexity for identifying the best arm.

\subsection{Synthetic Experiments}
\label{synthetic}
\begin{figure*}[t]
\centering
\subfloat[Average regret and stopping power in low SNR.\label{synthetic_low_snr}]{
\includegraphics[width=0.245\linewidth]{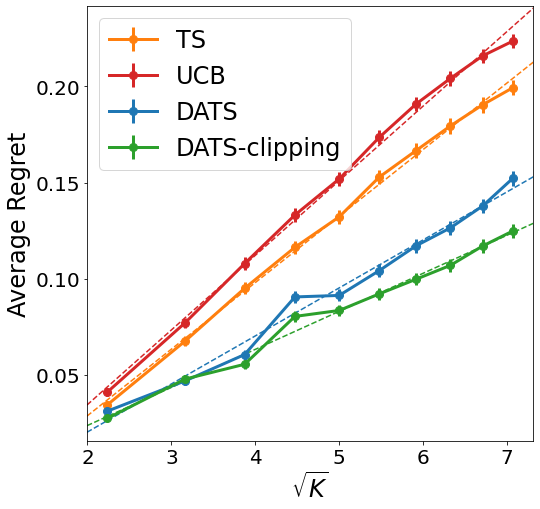}
\includegraphics[width=0.245\linewidth]{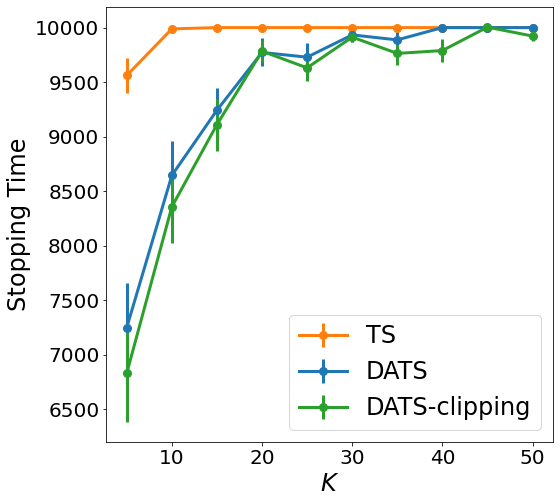}
}
\subfloat[Average regret and stopping power in high SNR.\label{synthetic_high_snr}]{
\includegraphics[width=0.24\linewidth]{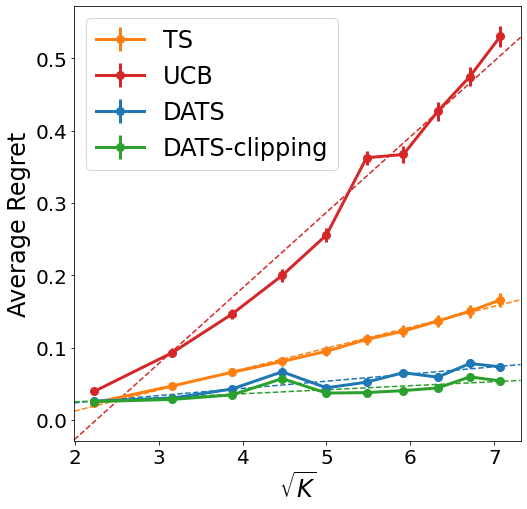}
\includegraphics[width=0.245\linewidth]{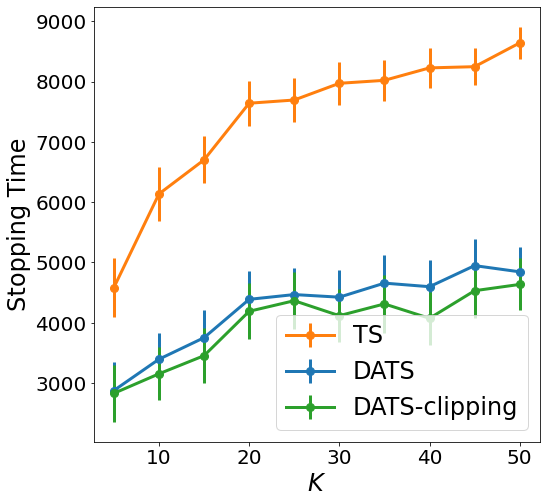}
}
\caption{Comparison of TS, UCB, DATS and DATS-clipping in terms of average regret vs. $\sqrt{K}$ and stopping power vs. $K$ (for probabilistic algorithms) in 20 synthetic domains (10 arm settings $K=5,...,50$ and 2 noise settings). Each domain is simulated 64 times and 1 standard error is shown.}
\label{fig:synthetic}
\end{figure*}

First, we present results on synthetic domains with varying number of arms and noise levels, as commonly done in literature \citep{russo2016simple}.
We simulate 20 synthetic domains with $K=5, 10, 15, 20, 25, 30, 35, 40, 45, 50$ where the true mean reward $\mu_a$ of each arm $a$ is drawn randomly from $\mathcal{N}(0, 0.125)$ in the low signal-to-noise ratio (SNR) setting and from $\mathcal{N}(0, 0.5)$ in the high SNR setting, while the rewards $r_{t, a}$ are drawn from $\mathcal{N}(\mu_a, 1)$. The multi-armed bandits run for horizon $T=10000$.  
For TS-Normal \citep{lattimore2020bandit}, described in section \ref{baselines}, we select a weak prior $\hat{\mu}_{0,a} = 0$, $\hat{\sigma}_{0, a}^2 = 10^6$ for all arms $a \in \cA$ and known noise variance $\sigma = 1$.
For UCB-Normal \citep{auer2002finite}, described in section \ref{baselines}, we tune $\beta$ among values $1, 1.5, 2, 2.5, 3, 4$ and select the one with the best performance in each domain. For DATS, we select $\gamma=0.01$. Finally, we also evaluate a simplified version of DATS, called DATS-clipping. Instead of arm elimination and uniform exploration among the non-eliminated arms that DATS introduces to control diminishing propensity scores, DATS-clipping applies TS to a modified ADR estimator which replaces $\pi_{s, a}$ in $\Gamma_{s, a}$ and in the adaptive weights $w_{s, a}$ by the clipped propensity $\max(\gamma, \pi_{s, a})$. Propensity score clipping is a common practice in the offline evaluation literature \citep{crump2009dealing}, but this modified estimator is no-longer unbiased. DATS-clipping uses $\gamma=0.001$.

We compare TS, UCB, DATS and DATS-clipping in terms of average regret, $\frac{1}{T}\sum_{t=1}^T(\mu_{a^*} - \mu_{a_t})$, as a function of the square root number of arms, $\sqrt{K}$, shown in the left subplots of \ref{synthetic_low_snr} and \ref{synthetic_high_snr} for the low SNR and the high SNR setting respectively.
We also compare the probabilistic bandits--TS, DATS, DATS-clipping--in terms of sample complexity for identifying the best arm with error tolerance $\delta = 0.05$ (i.e., with 95\% confidence) using the Bayesian stopping criterion, which stops at the first time $t^\dagger$ when there is an arm $a^\dagger$ such that its propensity score $\pi_{t^\dagger, a^\dagger}$ computed from its posterior $\bP(\mu_{a^\dagger} \in \cdot | \cH_{t^\dagger-1})$ satisfies $\pi_{t^\dagger, a^\dagger} \geq 1-\delta$ \citep{russo2016simple}. The right subplots of of \ref{synthetic_low_snr} and \ref{synthetic_high_snr} show this stopping time $t^\dagger$ for each algorithm as a function of the number of arms $K$
for the low SNR and the high SNR setting respectively.
Each one of the 20 domains is simulated over 64 simulations and the plots in Figure \ref{fig:synthetic} show the mean and one standard error for both the average regret and the stopping power metric. Both DATS and the DATS-clipping heuristic attain significantly better performance both in terms of average regret (linear dependence on $\sqrt{K}$ with smaller slope than TS and UCB) and in terms of stopping power (identify the best arm with 95\% confidence); DATS and DATS-clipping have better sample complexity than TS in all high SNR domains and in the low SNR domains with small $K$, while being within error from one-another.

\subsection{Semi-Synthetic Experiment Based on A/B Test Data}
\label{semisynthetic}
\begin{figure*}[t]
\centering
\subfloat[High SNR cumulative regret.]{\includegraphics[width=0.32\linewidth]{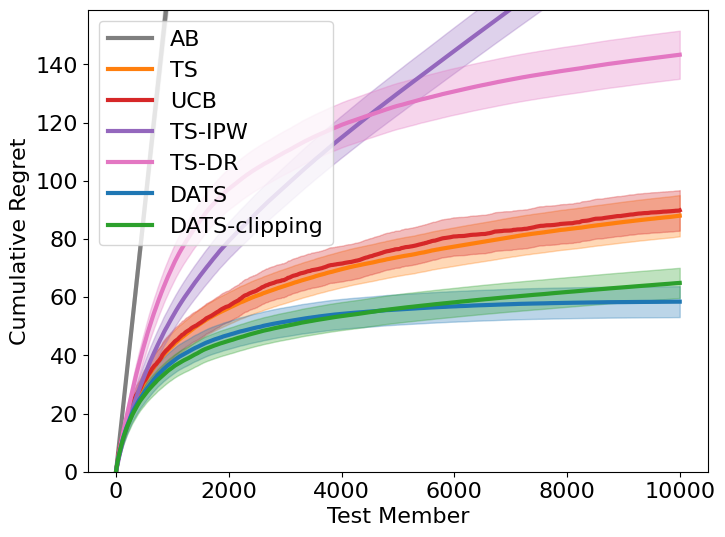}}
\hfill
\subfloat[Medium SNR cumulative regret.] {\includegraphics[width=0.32\linewidth]{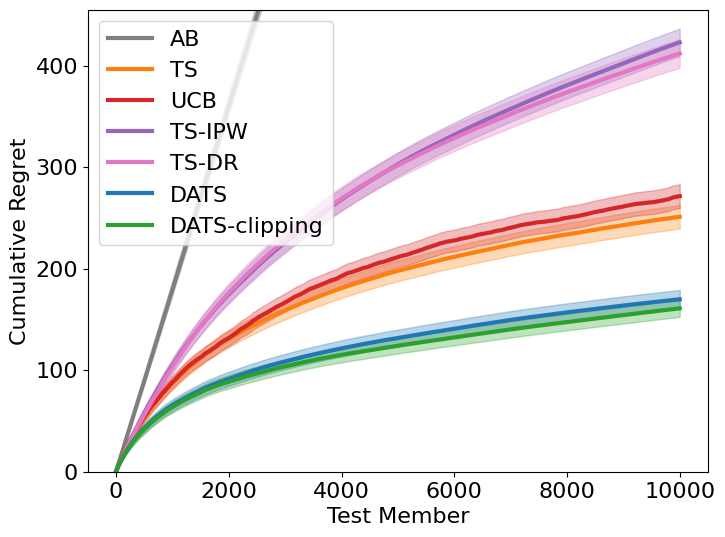}}
\hfill
\subfloat[Low SNR cumulative regret.]{\includegraphics[width=0.32\linewidth]{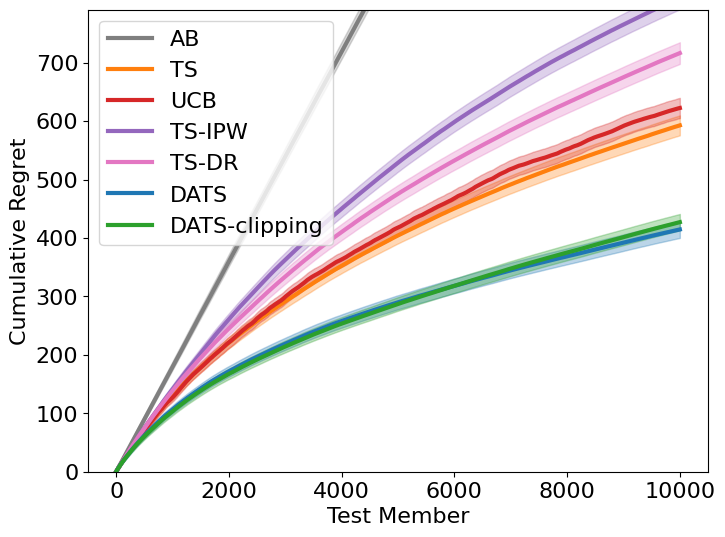}}
\\
\subfloat[High SNR stopping time.]{\includegraphics[width=0.32\linewidth]{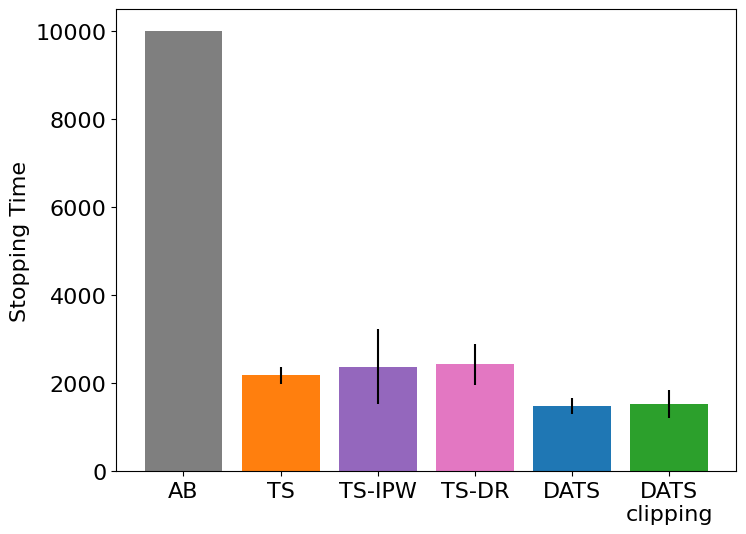}}
\hfill
\subfloat[Medium SNR stopping time.] {\includegraphics[width=0.32\linewidth]{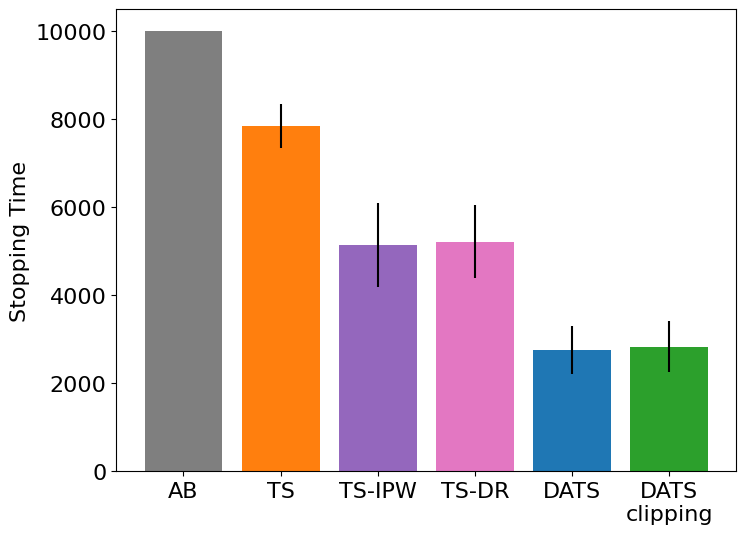}}
\hfill
\subfloat[Low SNR stopping time.]{\includegraphics[width=0.32\linewidth]{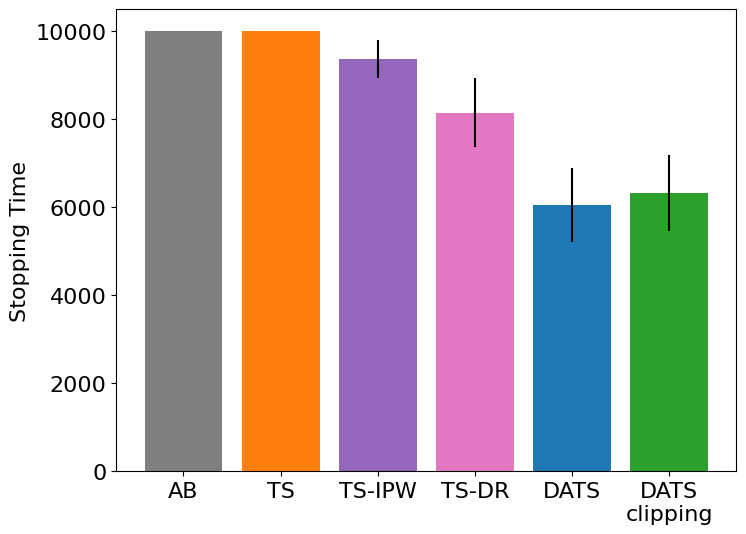}}
\caption{Comparison of A/B test, TS, UCB, TS-IPW, TS-DR, DATS and DATS-clipping in terms of cumulative regret and 95\% confidence stopping time (for stochastic algorithms) in high SNR ($\sigma = 0.32$), medium SNR ($\sigma = 0.64$) and low SNR ($\sigma = 1.28$) with $K = 6$ arms. The true reward means and medium SNR $\sigma$ estimated by a web-service's A/B test data. Bandits run for horizon $T = 10000$. Results are averaged over 64 simulations and 95\% confidence intervals are shown.}
\label{fig:semisynthetic}
\end{figure*}

We now evaluate the performance of DATS compared to baselines using data from an A/B test with 6 user-interface (UI) test cells that a web-service ran on its member base. For each test member that took part in this A/B test, the web-service selected uniformly at random one of the 6 UI cells, presented it to the member throughout the duration of the A/B test and collected measures of engagement of the member with the web-service, from which a target reward was defined. From the member-level data collected by the end of the A/B test, we extracted the sample mean of the target reward corresponding to each UI cell, $\mu =  [0, -0.05,  0.15,  0.02,  0.28,  0.2]$, and the sample standard deviation of the target reward across the 6 UI cells, $\sigma=0.64$. For reproducibility purposes, we use $\mu$ and $\sigma$ directly to simulate a multi-armed bandit domain with $K=6$ arms and $T=10000$ observations, where rewards are generated from $\mathcal{N}(\mu, \sigma^2 \text{I})$ (I is the $6\times6$ identity matrix).\footnote{The true mean reward $\mu_a$ of each arm $a$ and the noise $\sigma$ in this experiment are grounded in real data, hence its characterization as semi-synthetic (unlike the Section \ref{synthetic} where $\mu_a$ is generated at random for each arm $a$.)}. We simulate 3 noise domains: medium SNR with $\sigma = 0.64$ (observed in the A/B test); high SNR with $\sigma = 0.32$ (halving  the observed noise standard deviation) and low SNR with $\sigma = 1.28$ (doubling the observed noise standard deviation).

We evaluate TS ($\hat{\mu}_{0,a} = 0$, $\hat{\sigma}_{0, a}^2 = 10^6$, oracle $\sigma$), UCB ($\beta = 1, 1.5, 2, 2.5, 3, 4$), DATS ($\gamma = 0.01$) and DATS-clipping ($\gamma = 0.001$), as in section \ref{synthetic}. Additionally, we benchmark DATS against two more TS-based algorithms--TS-IPW and TS-DR--that replace the biased sample mean with the unbiased but not adaptively-weighted estimators $\hat{Q}^\text{IPW}$ and $\hat{Q}^\text{DR}$ respectively (more details in section \ref{subsec:motivation}) instead of $\hat{Q}^\text{ADR}$ as in DATS. The rest of the TS-IPW and TS-DR is implemented as in DATS (Algorithm \ref{alg:DATS}) with $\gamma = 0.01$\footnote{Section \ref{gamma-tuning} of the Supplemental Material explores tuning $\gamma$ for TS-IPW, TS-DR, DATS and DATS-clipping.}.

Figure \ref{fig:semisynthetic} shows that DATS and DATS-clipping have a clear advantage over baselines both in terms of cumulative regret and in terms of sample complexity for identifying the best arm with 95\% confidence in all three noise settings. Interestingly, TS-IPW and TS-DR, although unbiased, have worse regret performance than both TS and UCB, which use the biased sample mean. TS and best-tuned UCB are within error from each other. This indicates that using estimators which correct the sample mean bias stemming from the adaptive data collection is not enough if special attention is not given to the variance properties of these estimators. Both TS-IPW and TS-DR suffer from high variance due to the propensity scores in the denominator of their estimator which become small in an online multi-armed bandit setting\footnote{TS-DR has better variance properties than TS-IPW, and thus a bit better performance in the online setting.}. Through the variance-stabilizing property of the adaptive weights, DATS is unbiased with reasonably controlled variance and is able to provide a better confidence region for all arms' values, thus improving learning in the online setting.

\section{Finite-Time Regret Guarantees}
\label{sec:theory}

Our main theoretical result concerns an upper bound of the finite-time 
regret resulting from Algorithm~\ref{alg:DATS}. For notational simplicity,
we present the proof of the two-arm setting in this section, whereas the 
generalization to the $K$-arm case is deferred to the supplemental material. 

Theorem~\ref{thm:exp_regret_two} formally shows that for the two-arm
stochastic bandit problem the expected regret resulting from 
Algorithm~\ref{alg:DATS} matches the minimax lower bound $\Omega(\sqrt{T})$ 
(up to logarithmic factors). Throughout, we assume that the standard
deviation of the noise is $\sigma$, and that there exists $M>0$ such that $|r_t(a)|\le M$ 
for $t\in[T]$ and $a\in \cA$.

\begin{theorem}\label{thm:exp_regret_two}
For the two-arm stochastic bandit problem, the expected regret incurred 
by Algorithm~\ref{alg:DATS} is bounded as $\bE[R_T] \le O\big(\sqrt{T\log{T}}\big)$.
\end{theorem}

Without loss of generality, let $\cA = \{1,2\}$ and arm $1$ be the optimal arm;
let $\Delta = \mu_1 - \mu_2$ denote the gap between the two arms.
The proof of Theorem~\ref{thm:exp_regret_two} starts from establishing
that the estimator $\hat{\mu}_{t,a}$ concentrates around the true mean
$\mu_a$ with high probability; since $\hat{\mu}_{t,a}$ is a good estimator
for $\mu_a$, we claim that we are able to recognize the optimal arm fast
enough---so that we do not pay too much in the exploration phase---and 
we are committed to the optimal arm afterwards.

To start, we define $M_{t,a} := \sum^t_{s=1}\sqrt{\pi_{s,a}} \cdot
    \big(\hat{\Gamma}_{s,a}-\mu_{a} \big)$.
By construction, $\{M_{s,a}\}_{s \ge 1}$ is a martingale w.r.t.~the filtration 
$\{\calH_s\}_{s\ge 1}$. Let $\tau(a) := \min\{t: p_{t+1,a} < 1/T\}$. 
Since $p_{t+1,a}$ is measurable w.r.t.~$\calH_{t}$ 
for any $t\in[T]$, $\tau(a)$ is a stopping time w.r.t.~the filtration $\{\calH_t\}_{t \ge 1}$.
Consequently, $S_{t,a}:=M_{\tau(a)\wedge t,a}$ is a (stopped) martingale w.r.t.~$\{\calH_t\}_{t \ge 1}$.
For any $a\in \cA$, $t\in[T]$, the martingale difference of $S_{t,a}$ is given by  
\begin{align*}
    D_{t,a} = &S_{t,a} - S_{t-1,a} = M_{\tau(a) \wedge t,a} - M_{\tau(a)\wedge(t-1),a}
    = \indc{\tau(a) \ge t} \cdot \sqrt{\pi_{t,a}} \cdot \big(\hat{\Gamma}_{t,a} - \mu_a \big).
\end{align*}
On the event $\{\tau(a) \ge t\}$, the absolute value of the martingale difference $|D_{t,a}|$
can be bounded as,
\begin{align*}
    \sqrt{\pi_{t,a}} \cdot |\hat{\Gamma}_{t,a} - \mu_a| 
     = \sqrt{\pi_{t,a}} \cdot 
    \Big|\bar{r}_{t-1,a} -\mu_a + \frac{\indc{A_t = a}}{\pi_{t,a}} \cdot 
    (r_t - \bar{r}_{t-1,a})\Big|
     \le 2M + \frac{2M}{\sqrt{\gamma}}.
\end{align*}
Consequently,
$\sum^t_{s=1}|D_{s,t}|^2\le 4M^2\cdot(1+1/{\sqrt{\gamma}})^2\cdot t$. On the other hand,
\begin{align*}
    \sum^t_{s=1} \bE\big(D_{s,a}^2 \mid \calH_{s-1}\big) 
    =  \sum^t_{s=1} \indc{\tau(a) \ge s} \cdot \big[(1-\pi_{s,a})\cdot(\bar{r}_{s,a} - \mu_a)^2 
    + \sigma^2\big]
    \le (4M^2 + \sigma^2)\cdot t.
\end{align*}
We then make use of the following lemma to establish the concentration result.
\begin{lemma}[Theorem 3.26 of \citet{bercu2015concentration}]\label{lem:mg_concentration}
Let $\{M_n\}_{n \ge 1}$ be a square integrable martingale w.r.t.~the filtration $\{\mathcal{F}_t\}$
such that $M_0 = 0$. Then, for any positive $x$ and $y$,
\begin{align*}
    \bP \Big(M_n \ge x,~\sum^n_{i=1}(M_i - M_{i-1})^2 + 
    \sum^n_{i=1}\bE\big[(M_i - M_{i-1})^2 \mid \mathcal{F}_{i-1}\big] \le y \Big)
    \le \exp \Big(-\dfrac{x^2}{2y} \Big).
\end{align*}
\end{lemma}
Applying Lemma~\ref{lem:mg_concentration} to $S_{t,a}$, we have for any $x > 0$,
\begin{align*}
    \bP \Big(S_{t,a} \ge x \Big) 
    &= \bP\bigg(S_{t,a} \ge x, \sum^t_{s=1} D_{s,a}^2 
     + \sum^t_{s=1}\bE[D^2_{s,a}\mid \calH_{s-1}] \le
     \Big(4M^2\cdot \big(2 + {2}/{\sqrt{\gamma}} 
    + {1}/{\gamma}\big) + \sigma^2\Big)\cdot t \bigg)\\
    \le &\exp\Big(-\dfrac{x^2}{\big(8M^2 \cdot (2 + 2/\sqrt{\gamma} + 1/\gamma) + 2\sigma^2\big)\cdot t }  \Big).
\end{align*}
We now consider a ``good'' $E$ event on which the optimal arm is identified within 
$T_0 = O\big(\log(T)/\Delta^2\big)$ pulls, i.e., $E : = \{\tau(2) \le T_0\}$.
Lemma~\ref{lem:good_event}, whose proof is deferred to Supplementary Section~\ref{sec:proof_good_event},
guarantees the good event happening with high probability.
\begin{lemma}\label{lem:good_event}
The event $E$ happens with probability at least $1-2/T^2 - 2\log(T)/T^2$.
\end{lemma}
On the good event $E$, the sub-optimal arm is pulled for at most $T_0$ times, thus incurring a
regret at most $T_0 \cdot \Delta$. We can decompose the expected regret as,
\begin{align}
    \label{eq:problem_dependent_bnd}
    \bE\big[\text{Regret}(T,\pi)\big]&= \bE\big[\text{Regret}(T,\pi) \cdot \indc{E}\big] 
    + \bE\big[\text{Regret}(T,\pi) \cdot \indc{E^c}\big]\nonumber\\
    &\le  T_0\Delta + TM\cdot \dfrac{2}{T^2} + T + TM \cdot \dfrac{2\log(T)}{T^2} =  c(M,\gamma) \cdot \dfrac{\log(T)}{\Delta}
\end{align}
where $c(M,\gamma)$ is a constant that only depends on $M$ and $\gamma$. 
\eqref{eq:problem_dependent_bnd} provides a problem-dependent bound. To see
the problem-independent bound, note that
\begin{align*}
    \bE\big[\text{Regret}(T,\pi)\big] 
    &= \mathbf{1}\Big\{\Delta < \dfrac{\sqrt{c(M,\gamma)\log(T)}}{\sqrt{T}}\Big\}
    \cdot \bE\big[\text{Regret}(T,\pi)\big]\\
    &+ \mathbf{1} \Big\{\Delta \ge \dfrac{\sqrt{c(M,\gamma)\log(T)}}{\sqrt{T}}\Big\}
    \bE\Big[\text{Regret}(T,\pi)\Big]
    \le \sqrt{c(M,\gamma)T\log(T)}.
\end{align*}
As a by-product of the proof of Theorem~\ref{thm:exp_regret_two}, we obtain 
a high probability regret bound---this is stated in Corollary~\ref{cor:high_prob_regret}.
\begin{corollary}\label{cor:high_prob_regret}
In the setting of Theorem~\ref{thm:exp_regret_two}, with probability 
$\geq 1-2/T^2 - 2\log(T)/T^2$, the regret can be bounded as $ \text{\textnormal{Regret}}(T, \pi)  = O\big(\sqrt{T\log(T)}\big)$.
\end{corollary}
Finally, we state the result for the general $K$-arm case, the proof 
of which can be found in Supplementary Section~\ref{sec:proof_karm}.

\begin{theorem}\label{thm:regret_karm}
For the $K$-arm stochastic bandit problem, the expected regret incurred 
by Algorithm~\ref{alg:DATS} is bounded as $\bE\big[\regret(T,\pi)\big] \le O\big(K^2 \cdot \sqrt{T\log{T}}\big)$.
\end{theorem}

\section{Conclusion}
\label{conclusion}

The adaptive data collection in a MAB setting poses an inference challenge that is overlooked in traditional and provably optimal algorithms such as UCB and TS.
In this paper, we proposed doubly-adaptive Thompson sampling (DATS), which incorporated advances in (offline) causal inference estimators for adaptive experimental design and modified them appropriately in order to render them suitable for the exploration/exploitation trade-off present in (online) MABs. Beyond DATS, the aim for this paper was to motivate a general approach for designing online learning algorithms with more accurate estimation methods
that accounts for the biases stemming from the data-collecting mechanism.

\section{Societal Impact}
\label{societal}
Experimentation consumes resources and hence are expensive in the real-world.
As such, experimentation methods that achieve better efficiency with less experimentation capacity would be the desired blueprint for the next-generation applications in our AI-powered society. 
Adaptive experimentation holds great promise over traditional A/B tests for real-world domains (e.g. web-service testing, clinical trials etc.), as it aims to maximize the test population's reward during learning and it is effective in reducing the test duration by directing test-traffic into actions that are more promising rather than splitting traffic equally among arms. In this work, we advance the field of adaptive experimentation by introducing the idea of reliable inference \textit{during} online decision making. We design a multi-armed bandit algorithm, which is shown empirically to improve the test population reward and decrease the test samples required to identify the best arm, while attaining the optimal worst-case regret bound guarantees in terms of time horizon (proven theoretically) and in terms of number of arms (shown empirically). 
At the same time, in our algorithm, like in any multi-armed bandit algorithm that explores all arms to some degree or in an A/B test, the decision maker needs to consider the safety of all arms to which the test population is exposed and inform the test population accordingly. Also, multi-armed bandit algorithms, including the one this paper proposes, focus on maximizing the reward over the entire test population at the potential expense of some individual-level reward. An in-depth study of such considerations requires future work. 

\bibliography{main}
\bibliographystyle{apalike}


\newpage
\appendix
\input{supplement}

\end{document}

%% file: supplement.tex
\begin{center}
\large Supplemental Material for:
\\[1ex]{\Large\bf Online Multi-Armed Bandits with Adaptive Inference}
\\[1ex]\large Maria Dimakopoulou, Zhimei Ren, Zhengyuan Zhou
\end{center}

\section{Supplementary Proofs}\label{sec:proof}

\subsection{Proof of Lemma~\ref{lem:good_event}}
\label{sec:proof_good_event}
To start, we define
\begin{align*}
    T_0 = \frac{4}{\Delta^2\gamma} \cdot \Big( 2\kappa M \cdot(1+1/\sqrt{\gamma})
    \cdot\sqrt{2\log(T) - \log\log(T)} + 2M\sqrt{(4+
    4/\sqrt{\gamma}+2/\gamma + 1)}\cdot \log(T) \Big)^2. 
\end{align*}
The probability of the good event $E$ not happening is
\begin{align*}
    \bP(E^c) = \bP\big( \tau(2)> T_0\big) 
    = \bP\big(\tau(1) \le T_0, \tau(2) > T_0 \big) 
    + \bP\big(\tau(1) > T_0, \tau(2) > T_0 \big).
\end{align*}
The first term corresponds to the probability of eliminating the optimal arm
within the first $T_0$ pulls:
\begin{align}\label{eq:term1}
    \bP(\tau(1) \le T_0,\tau(2) > T_0) = \sum^{T_0}_{t=1}\bP\big(\tau(1) = t, \tau(2) > T_0 \big)
\end{align}
Note that  
\begin{align*}
    \tau(1) = t \Longrightarrow ~ &\bP\big(\Tilde{r}_{t+1,1} > \Tilde{r}_{t+1,2} \mid \calH_t\big) < \frac{1}{T}\\
    \iff & \bP\bigg(\dfrac{\Tilde{r}_{t+1,1} - \hat{\mu}_{t,1}- \Tilde{r}_{t+1,2} + \hat{\mu}_{t,2}}
    {\kappa\sqrt{\hat{\sigma}^2_{t,1} + \hat{\sigma}^2_{t,2}}} > \dfrac{\hat{\mu}_{t,2} - \hat{\mu}_{t,1}}
    {\kappa\sqrt{\hat{\sigma}^2_{t,1} + \hat{\sigma}^2_{t,2}}} \, \Big| \, \calH_t\bigg) < \frac{1}{T}\\
    \iff & 1 - \Phi\bigg(\dfrac{\hat{\mu}_{t,2} - \hat{\mu}_{t,1}}{\kappa\sqrt{\hat{\sigma}^2_{t,1} 
    + \hat{\sigma}^2_{t,2}}} \bigg) < \frac{1}{T} \\
    \Rightarrow & \dfrac{\hat{\mu}_{t,2} - \hat{\mu}_{t,1}}{\kappa\sqrt{\hat{\sigma}^2_{t,1}
    + \hat{\sigma}^2_{t,2}}} > \sqrt{2\log(T) - \log \log(T)}.
\end{align*}
Consequently,
\begin{align}\label{eq:term1_a}
    \eqref{eq:term1} \le &\sum^{T_0}_{t=1} \bP\Big(\tau(1) = t, \tau(2) > T_0, \hat{\mu}_{t,2} - \hat{\mu}_{t,1}
    > \kappa \cdot \sqrt{2\log(T) - \log \log(T)} \cdot \sqrt{\hat{\sigma}^2_{t,1} + \hat{\sigma}^2_{t,2}}\Big)\nonumber\\
    \le & \sum^{T_0}_{t=1} \bP\Big(\tau(1) = t, \tau(2) > T_0, \hat{\mu}_{t,2} - \mu_2 - \hat{\mu}_{t,1} + \mu_1 >
    \Delta + \kappa \cdot \sqrt{2\log(T) - \log \log(T)}\cdot \sqrt{\hat{\sigma}^2_{t,1} + \hat{\sigma}^2_{t,2}} \Big)\nonumber\\
    \le & \sum^{T_0}_{t=1} \bP\Big(\tau(1) = t, \tau(2) > T_0, \hat{\mu}_{t,2} - \mu_2 >  \dfrac{\Delta}{2} +
    \dfrac{\kappa\sqrt{2\log(T) - \log\log(T)}}{2}\cdot\sqrt{\hat{\sigma}^2_{t,1} + \hat{\sigma}^2_{t,2}}\Big)\nonumber\\
     &\qquad + \bP\Big(\tau(1) = t, \tau(2) > T_0, -\hat{\mu}_{t,1} + \mu_1 > \dfrac{\Delta}{2}
    + \dfrac{\kappa\sqrt{\log(T) - \log\log(T)}}{2} \cdot \sqrt{\hat{\sigma}^2_{t,1} + \hat{\sigma}^2_{t,2}}\Big)\nonumber\\
    \le & \sum^{T_0}_{t=1} \bP\left(\tau(1) = t, \tau(2) > T_0, M_{t,t}(2) > \frac{\Delta}{2} \sum^t_{s=1}\sqrt{\pi_{s,2}} + 
    \dfrac{\kappa\hat{\sigma}_{t,2}\cdot\sqrt{2\log(T)-\log\log(T)}}{2}\cdot \sum^t_{s=1}\sqrt{\pi_{s,a}}
    \right) \nonumber\\
    & \qquad+ \bP\bigg(\tau(1) = t, \tau(2) > T_0, -M_{t,t}(1) > \frac{\Delta}{2}\sum^t_{s=1}\sqrt{\pi_{s,1}} + 
    \dfrac{\kappa\hat{\sigma}_{t,1} \cdot \sqrt{2\log(T) - \log \log(T)} }{2} \cdot \sum^t_{s=1}\sqrt{\pi_{s,a}}\bigg) 
\end{align}
For any $a\in\{1,2\}$ and $t\in [T_0]$, on the event $\{\tau(1) = t, \tau(2) > T_0\}$, $S_{t,a} = M_{t,a}$, and 
the estimated variance can be bounded as,
\begin{align*}
    \hat{\sigma}^2_{t,a} \cdot \Big(\sum^t_{s=1}\sqrt{\pi_{s,a}}\Big)^2 =
    \sum^t_{s=1} \pi_{s,a} \cdot \Big[ \big(\hat{\Gamma}_{s,a} - \hat{\mu}_{s,a}\big)^2 +1 \Big] 
    \ge \gamma t.
\end{align*}
As a result,
\begin{align*}
    \eqref{eq:term1_a} \le & \sum^{T_0}_{t=1} \bP\left( \tau(1) = t, \tau(2) > t_0,
     S_{t,1} > \dfrac{\Delta\sqrt{\gamma}t}{2} +
      \dfrac{\kappa\sqrt{\gamma t} \cdot\sqrt{2\log(T) - \log \log(T)}}{2}\right) \\
     + & \bP\left( \tau(1) = t, \tau(2) > t_0,
     S_{t,2} > \dfrac{\Delta\sqrt{\gamma}t}{2} +  
      \dfrac{\kappa \sqrt{\gamma t} \cdot  \sqrt{2\log(T) - \log \log(T)}}{2}\right)\\ 
     \le & 2\sum^{T_0}_{t = 1} \exp\Big(- \dfrac{\kappa \gamma (2\log(T) - \log \log (T))}
     {48M^2 (2 + 2/\sqrt{\gamma} + 1 / \gamma) + 12\sigma^2}\Big)\\
     \le & \dfrac{2T_0 \log(T) }{T^3} \le \dfrac{2\log(T)}{T^2}, 
\end{align*}
where the last inequality is due to the choice $\kappa = \sqrt{(48M^2(2 + 2/\sqrt{\gamma}+1) + 12M^2) / \gamma}$.
We now proceed to bound $\bP(\tau(1) > T_0, \tau(2) > T_0)$:
\begin{align}\label{eq:term2}
    &\bP \Big(\tau(1) > T_0, \tau(2) >T_0\Big)\nonumber\\
    = &\bP \Big(\tau(1) > T_0, \tau(2) > T_0,
    \hat{\mu}_{T_0,2} - \hat{\mu}_{T_0,1} \ge \kappa \sqrt{2\log(T) - 2\log \log(T)}
    \sqrt{\hat{\sigma}_{T_0,1}^2 + \hat{\sigma}^2_{T_0,2}} \Big)\nonumber\\
    \le & \bP\Big(\tau(1) > T_0, \tau(2) > T_0,  \hat{\mu}_{T_0,2} - \mu_2 
    \ge \dfrac{\Delta}{2} + \dfrac{\kappa \sqrt{\log(T) - \log\log{T}}}{2}
    \sqrt{\hat{\sigma}^2_{T_0,1} + \hat{\sigma}^2_{T_0,2}}\Big)\nonumber\\
    & \quad + \bP\Big(\tau(1) > T_0, \tau(2) > T_0,  \mu_1 - \hat{\mu}_{T_0,1} 
    \ge \dfrac{\Delta}{2} + \dfrac{\kappa\sqrt{2\log(T) - \log\log(T)}}{2}
    \sqrt{\hat{\sigma}^2_{T_0,1} + \hat{\sigma}^2_{T_0,2}}\Big)\nonumber\\
    \le &\bP\Big(S_{T_0,1} \ge \dfrac{\Delta}{2}\sum^{T_0}_{s=1}\sqrt{\pi_{s,1}} + 
    \dfrac{\kappa \sqrt{2\log(T) - \log\log(T)}}{2}
    \sqrt{\hat{\sigma}^2_{T_0,1} + \hat{\sigma}^2_{T_0,2}}
    \sum^{T_0}_{s=1}\sqrt{\pi_{s,1}}\Big) \nonumber\\ 
    & \quad + \bP\Big(-S_{T_0,2} \ge \dfrac{\Delta}{2}\sum^{T_0}_{s=1}\sqrt{\pi_{s,2}} + 
    \dfrac{\kappa \sqrt{2\log(T) - \log\log(T)}}{2}
    \sqrt{\hat{\sigma}^2_{T_0,1} + \hat{\sigma}^2_{T_0,2}}\sum^{T_0}_{s=1}\sqrt{\pi_{s,2}}\Big).
\end{align}
With the choice of $T_0$,
\begin{align*}
    \eqref{eq:term2} \le & 2 \cdot \bP \Big(|S_{T_0,1}|\ge 
    \sqrt{  T_0\log{T}\cdot \big(16M^2(2+2/\sqrt{\gamma}+1/\gamma)+4M^2\big)}
    \Big)
    \le 2T^{-2}.
\end{align*}
Combining everything, we conclude that $\bP(E) \ge 1-2T^{-2} - T^{-2} \cdot \log{T}$.

\subsection{Proof of Theorem~\ref{thm:regret_karm}}\label{sec:proof_karm}
Without loss of generality, we let arm $1$ be the optimal arm, and define the suboptimality gap for each $a\in \{2,\ldots,K\}$:
\begin{align*}
    \Delta_a := \mu_1 - \mu_a. 
\end{align*}
Similar to the two-arm case, we define for each $a \in \{2,\ldots,K\}$ the stopping time
\begin{align*}
    \tau(a) = \min\{t: p_{t + 1, a} < 1/T\},
\end{align*}
where $p_{t+1, a} = \min_{a' \in \cA_t}\bP(\Tilde{r}_{t+1, a} > \Tilde{r}_{t+1, a'} \mid \calH_t)$.
We additionally define 
\begin{align*}
T_a \,:=\,\frac{4K\log{T}}{\Delta_a^2 \gamma} \cdot \big(8M\kappa + 2M\sqrt{2+2/\sqrt{K/\gamma}+K/\gamma}\big).
\end{align*}
For each $a\neq 1$, let $E_a := \{\tau(a) \le T_a, \tau(1) > T_a\}$ denote the good event. 
We now proceed to bound the probability of the good event not happening. To start, note that
\begin{align*}
    \bP(E_a^c) = \bP\big(\tau(1) \le T_a\big)
    + \bP\big(\tau(a) > T_a, \tau(1) > T_a\big).
\end{align*}
The first term can be decomposed as:
\begin{align}\label{eq:term1_k}
    \bP(\tau(1) \le T_a) = & \sum^{T_a}_{t=1}\bP(\tau(1) = t)
    \le \sum^{T_a}_{t=1}\bP(p_{t+1,1} < 1/T).
\end{align}
Above, $p_{t+1,1} < 1/T$ means that there exists $a' \neq 1$ such that $\tau(a') > t$ and 
$\bP(\Tilde{r}_{t+1,1} > \Tilde{t}_{t,a'} \mid \calH_t) < 1/T$. That is,
\begin{align}\label{eq:term1_karm_2}
    \eqref{eq:term1_k} \le & \sum^{T_a}_{t=1}\sum_{a'\neq 1} \bP(\tau(a') > t, \bP(\Tilde{r}_{t+1,1} > \Tilde{r}_{t+1,a'} \mid \calH_t) < 1/T)\nonumber\\
    \le &\sum^{T_a}_{t=1}\sum_{a'\neq 1} \bP\Big(\tau(a') > t, \hat{\mu}_{t,a'} - \hat{\mu}_{t,1} > \kappa \cdot \sqrt{2\log{T} - \log \log{T}}
    \cdot \sqrt{\hat{\sigma}^2_{t,1} + \hat{\sigma}^2_{t,a'}}\Big)\nonumber\\
    \le & \sum^{T_a}_{t=1}\sum_{a'\neq 1} \bP\Big(\tau(a') > t, \hat{\mu}_{t,a'} - \mu_{a'} > 
    \dfrac{\Delta_{a'}}{2} + \dfrac{\kappa }{2} \cdot \sqrt{2\log{T} - \log \log{T}}\cdot
    \sqrt{\hat{\sigma}^2_{t,1} + \hat{\sigma}^2_{t,a'}}\Big) \nonumber\\
     & \qquad + \bP\Big(\tau(a') > t, \mu_1 - \hat{\mu}_{t,1} > \dfrac{\Delta_{a'}}{2} + \dfrac{\kappa }{2} 
     \cdot \sqrt{2\log{T} - \log \log{T}}\cdot \sqrt{\hat{\sigma}^2_{t,1} + \hat{\sigma}^2_{t,a}}\Big)\nonumber\\
    \le & \sum^{T_a}_{t=1}\sum_{a'\neq a} \bP\Big(S_{t,a'} \ge \dfrac{\Delta_{a'}}{2} \cdot\sqrt{\dfrac{\gamma}{K}}\cdot t 
     + \dfrac{\kappa\sqrt{\gamma t}}{2\sqrt{K}}\cdot  \sqrt{2\log{T} - \log \log{T}}
     \Big)\nonumber\\
    & \qquad + \bP\Big(S_{t,1} \ge \dfrac{\Delta_{a'}}{2}\cdot \sqrt{\dfrac{\gamma}{K}}\cdot t + 
    \dfrac{\kappa\sqrt{\gamma t}}{2\sqrt{K}} \cdot \sqrt{2\log{T} - \log \log{T}}
    \Big).
\end{align}
Applying Lemma~\ref{lem:mg_concentration}, we have that
\begin{align*}
    \eqref{eq:term1_karm_2} \le & 2\sum^{T_a}_{t=1}\sum_{a'\neq 1}\exp \left(- \dfrac{\gamma \big(\Delta_{a'}\sqrt{t} 
    + \kappa\sqrt{2\log{T} - \log\log{T}})^2}{32M^2K(2+2\sqrt{K}/\sqrt{\gamma} + K/\gamma\big)}\right)\\
    \le & 2\sum^{T_a}_{t=1}\sum_{a'\neq 1}\exp\left(-\dfrac{\gamma \kappa^2  (2\log{T} - \log\log{T})}
    {32M^2 K(2+2\sqrt{K/\gamma}+K/\gamma)}\right)
    \le 2\dfrac{K T_a \log{T}}{T^3}\le \dfrac{2K\log{T}}{T^2},
\end{align*}
where the last inequality is due to $\kappa = \sqrt{48 M^2 K(2+2\sqrt{K/\gamma}+K/\gamma)/\gamma}$.
Switching to the second term, we have that
\begin{align}\label{eq:term2_k}
    &\bP\Big(\tau(a) > T_a, \tau(1) > T_a \Big) \nonumber\\
    \le &  \bP\Big(\tau(a) > T_a, \tau(1) > T_a, 
    \bP(\Tilde{r}_{T_a+1,a} > \Tilde{r}_{T_a+1,1} \mid \calH_{T_a}) > 1/T\Big)\nonumber\\
    \le & \bP\Big(\tau(a) > T_a, \tau(1) > T_a, \hat{\mu}_{T_a+1,a} - \hat{\mu}_{T_a+1,1}
    > - \kappa  \sqrt{\hat{\sigma}^2_{T_a,1} + \hat{\sigma}^2_{T_a,a}}\sqrt{2\log{T} - 2\log \log{T}}\Big)\nonumber\\
    \le & \bP\Big(\tau(a) > T_a, \tau(1) > T_a, \hat{\mu}_{T_a+1,a} - \mu_a
    > \dfrac{\Delta_a}{2} - \dfrac{\kappa }{2} \sqrt{\hat{\sigma}^2_{T_a,1} + \hat{\sigma}^2_{T_a,a}}\sqrt{2\log{T} - 2\log \log{T}}\Big)\nonumber\\
    & \qquad + \bP\Big(\tau(a) > T_a, \tau(1) > T_a, \mu_1 - \hat{\mu}_{t+1,1}
    >\dfrac{\Delta_a}{2} - \dfrac{\kappa }{2}\sqrt{\hat{\sigma}^2_{T_a,1} + \hat{\sigma}^2_{T_a,a}}\sqrt{2\log{T} - 2\log \log{T}}\Big)\nonumber\\
    \le & \bP\Big(S_{T_a,a} > \dfrac{\Delta_a T_a}{2}\sqrt{\dfrac{\gamma}{K}} -
    4 M \kappa \sqrt{T_a }\sqrt{2\log{T} - 2\log \log{T}} \Big)\nonumber\\
    & \qquad+ \bP\Big(- S_{T_a,1} > \dfrac{\Delta_a T_a}{2}\sqrt{\dfrac{\gamma}{K}} -
    4 M \kappa \sqrt{T_a }\sqrt{2\log{T} - 2\log \log{T}} \Big)
\end{align}
Applying Lemma~\ref{lem:mg_concentration} and with the choice of $T_a$, we arrive at $\eqref{eq:term2_k} \le  T^{-2}$.
Finally, denoting the number of pulls of arm $a$ by $N(a)$, we decompose the regret as
\begin{align*}
    \bE\big[R(T, \pi)\big] = &\sum^K_{a=2} \Delta_a \bE \big[N(a)\big]
    =  \sum^K_{a=1} \Delta_a \Big(\bE\big[N(a)E_a\big] + \bE\big[N(a)E_a^c\big]\Big)\\
    \le & \sum^K_{a=2} \Delta_a \Big( T_a + \dfrac{3MK \cdot \log{T}}{T} \Big)
    = O\big(K^2\cdot \log{(TK)}\big)\cdot \Big(\sum^K_{a=2} \dfrac{1}{\Delta_a}\Big) 
\end{align*}
And the problem-independent bound is 
\begin{align*}
    \bE\Big[R(T,\pi)\Big] = & \sum_{\Delta_a > K\sqrt{\log{T}}/\sqrt{T}} \Delta_a \bE[N(a)]
    + \sum_{\Delta_a \le K\sqrt{\log{T}}/\sqrt{T}} \Delta_a \bE[N(a)]\\
    =  &O\Big(K^2 \cdot \sqrt{T\log{T}}\Big). 
\end{align*}

\section{Additional Results for the Empirical Investigation}

\subsection{Sensitivity to propensity-controlling parameter $\gamma$}
\label{gamma-tuning}
\begin{figure*}[t]
\centering
\subfloat[Cumulative regret.]{\includegraphics[width=0.58\linewidth,,valign=t]{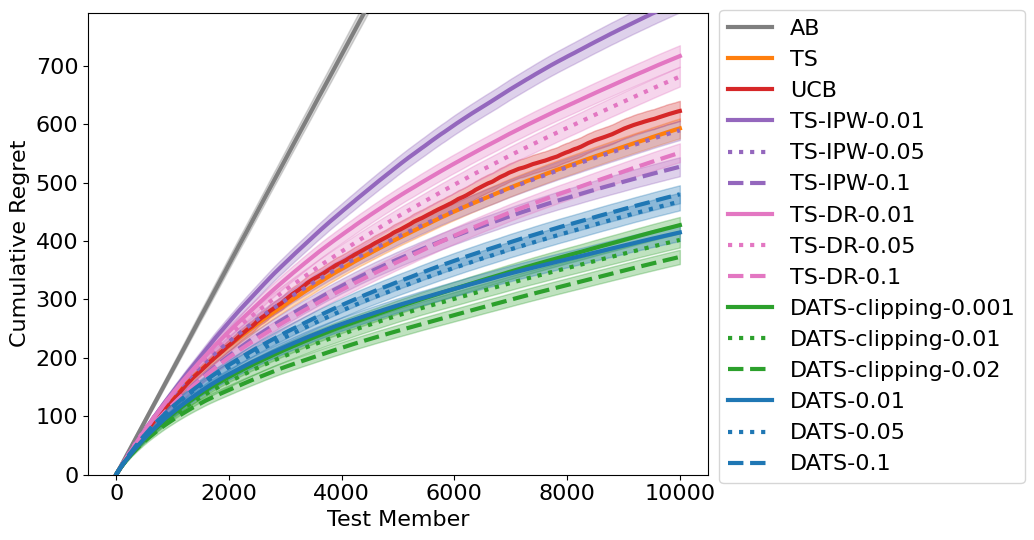}}
\hfill
\subfloat[Stopping power.]{\includegraphics[width=0.41\linewidth,valign=t]{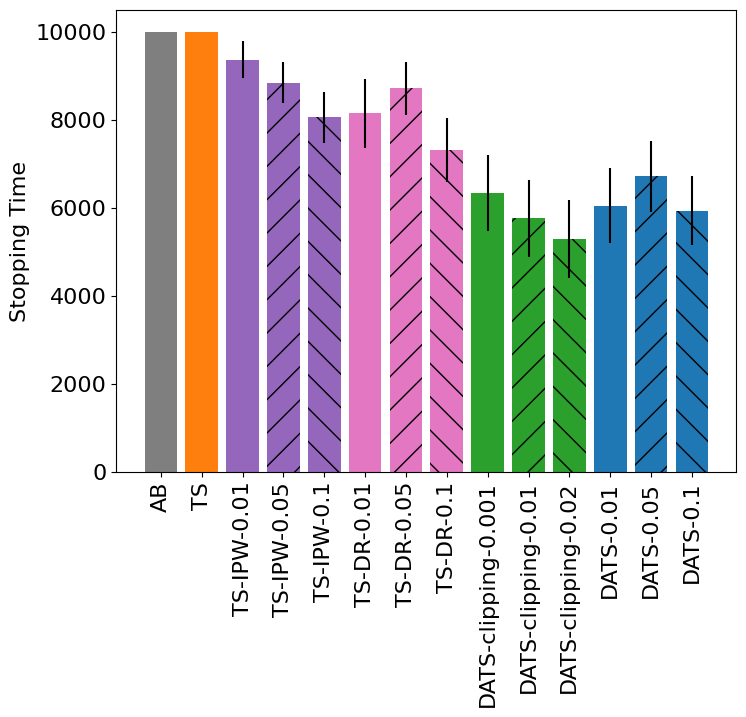}}
\caption{Sensitivity of TS-IPW, TS-DR, DATS and DATS-clipping to propensity-controlling parameter $\gamma$ in terms of cumulative regret and stopping power in the low SNR setting of Section's \ref{semisynthetic} semi-synthetic experiment}
\label{fig:gamma_tuning}
\end{figure*}

We use the low SNR setting of the semisynthetic experiment described in Section \ref{semisynthetic} of the main paper to explore the sensitivity of the ``causal'' TS algorithms which use unbiased estimators $\hat{Q}^\text{IPW}_{t, a}$ (TS-IPW), $\hat{Q}^\text{DR}_{t, a}$ (TS-DR) and $\hat{Q}^\text{ADR}_{t, a}$ (DATS) to the choice of parameter $\gamma$, which controls how small the propensity scores get. In the case of TS-IPW, TS-DR and TS-ADR, $\gamma$ is the level of uniform exploration among the non-eliminated arms (see Section \ref{algorithm} and Algorithm \ref{alg:DATS} for more details). In the case of the heuristic, DATS-clipping, $\gamma$ is merely a clipping threshold of the propensity score in the $\hat{Q}^\text{ADR}_{t, a}$ estimator (rendering it no-longer unbiased).
We try uniform-explore values $\gamma = 0.01, 0.05, 0.1$ for TS-IPW, TS-DR and DATS and clipping values $\gamma = 0.001, 0.01, 0.02$ for the heuristic DATS-clipping.
In Figure \ref{fig:gamma_tuning}, we see that both DATS and its heuristic DATS-clipping are robust to the choice of parameter $\gamma$, whereas TS-IPW and TS-DR which use the non-adaptively-weighted and high-variance estimators $\hat{Q}^\text{IPW}_{t, a}$ and $\hat{Q}^\text{DR}_{t, a}$ instead of $\hat{Q}^\text{ADR}_{t, a}$. We observe that by tuning $\gamma$, TS-IPW and TS-DR can improve their performance and even become competitive to TS and UCB, whereas with the default parameter TS-IPW and TS-DR were under-performed by TS and best-tuned UCB due to their poorly-controlled variance.
On the other hand, thanks to the variance stabilization properties of the adaptive weights, DATS and DATS-clipping remain the best performing variants even among the best-tuned TS-IPW and TS-DR.

%% file: main.bbl
\begin{thebibliography}{}

\bibitem[Agarwal et~al., 2014]{agarwal2014taming}
Agarwal, A., Hsu, D., Kale, S., Langford, J., Li, L., and Schapire, R. (2014).
\newblock Taming the monster: A fast and simple algorithm for contextual
  bandits.
\newblock In {\em International Conference on Machine Learning}, pages
  1638--1646. PMLR.

\bibitem[Agrawal, 1995]{agrawal_1995}
Agrawal, R. (1995).
\newblock Sample mean based index policies by o(log n) regret for the
  multi-armed bandit problem.
\newblock {\em Advances in Applied Probability}, 27(4):1054–1078.

\bibitem[Agrawal et~al., 2017]{agrawal2017thompson}
Agrawal, S., Avadhanula, V., Goyal, V., and Zeevi, A. (2017).
\newblock Thompson sampling for the mnl-bandit.
\newblock {\em arXiv preprint arXiv:1706.00977}.

\bibitem[Agrawal and Goyal, 2012]{agrawal2012analysis}
Agrawal, S. and Goyal, N. (2012).
\newblock Analysis of thompson sampling for the multi-armed bandit problem.
\newblock In {\em Conference on learning theory}, pages 39--1. JMLR Workshop
  and Conference Proceedings.

\bibitem[Agrawal and Goyal, 2013a]{agrawal2013further}
Agrawal, S. and Goyal, N. (2013a).
\newblock Further optimal regret bounds for thompson sampling.
\newblock In {\em Artificial intelligence and statistics}, pages 99--107. PMLR.

\bibitem[Agrawal and Goyal, 2013b]{AG2013a}
Agrawal, S. and Goyal, N. (2013b).
\newblock Further optimal regret bounds for thompson sampling.
\newblock In {\em Artificial intelligence and statistics}, pages 99--107.

\bibitem[Agrawal and Goyal, 2013c]{AG2013b}
Agrawal, S. and Goyal, N. (2013c).
\newblock Thompson sampling for contextual bandits with linear payoffs.
\newblock In {\em International Conference on Machine Learning}, pages
  127--135.

\bibitem[Agrawal and Goyal, 2017]{agrawal2017near}
Agrawal, S. and Goyal, N. (2017).
\newblock Near-optimal regret bounds for thompson sampling.
\newblock {\em Journal of the ACM (JACM)}, 64(5):1--24.

\bibitem[Audibert et~al., 2009]{audibert2009minimax}
Audibert, J.-Y., Bubeck, S., et~al. (2009).
\newblock Minimax policies for adversarial and stochastic bandits.
\newblock In {\em COLT}, volume~7, pages 1--122.

\bibitem[Auer, 2002]{auer2002using}
Auer, P. (2002).
\newblock Using confidence bounds for exploitation-exploration trade-offs.
\newblock {\em Journal of Machine Learning Research}, 3(Nov):397--422.

\bibitem[Auer et~al., 2002]{auer2002finite}
Auer, P., Cesa-Bianchi, N., and Fischer, P. (2002).
\newblock Finite-time analysis of the multiarmed bandit problem.
\newblock {\em Machine learning}, 47(2-3):235--256.

\bibitem[Bercu et~al., 2015]{bercu2015concentration}
Bercu, B., Delyon, B., and Rio, E. (2015).
\newblock {\em Concentration inequalities for sums and martingales}.
\newblock Springer.

\bibitem[Berry and Fristedt, 1985]{berry1985bandit}
Berry, D.~A. and Fristedt, B. (1985).
\newblock Bandit problems: sequential allocation of experiments (monographs on
  statistics and applied probability).

\bibitem[Bietti et~al., 2018]{bietti2018contextual}
Bietti, A., Agarwal, A., and Langford, J. (2018).
\newblock A contextual bandit bake-off.
\newblock {\em arXiv preprint arXiv:1802.04064}.

\bibitem[Bowden and Trippa, 2017]{bowden2017unbiased}
Bowden, J. and Trippa, L. (2017).
\newblock Unbiased estimation for response adaptive clinical trials.
\newblock {\em Statistical methods in medical research}, 26(5):2376--2388.

\bibitem[Bubeck and Cesa-Bianchi, 2012]{bubeck2012regret}
Bubeck, S. and Cesa-Bianchi, N. (2012).
\newblock Regret analysis of stochastic and nonstochastic multi-armed bandit
  problems.
\newblock {\em arXiv preprint arXiv:1204.5721}.

\bibitem[Carpentier et~al., 2011]{carpentier2011upper}
Carpentier, A., Lazaric, A., Ghavamzadeh, M., Munos, R., and Auer, P. (2011).
\newblock Upper-confidence-bound algorithms for active learning in multi-armed
  bandits.
\newblock In {\em International Conference on Algorithmic Learning Theory},
  pages 189--203. Springer.

\bibitem[Chapelle and Li, 2011]{chapelle2011empirical}
Chapelle, O. and Li, L. (2011).
\newblock An empirical evaluation of thompson sampling.
\newblock {\em Advances in neural information processing systems},
  24:2249--2257.

\bibitem[Chu et~al., 2011]{chu2011contextual}
Chu, W., Li, L., Reyzin, L., and Schapire, R. (2011).
\newblock Contextual bandits with linear payoff functions.
\newblock In {\em Proceedings of the Fourteenth International Conference on
  Artificial Intelligence and Statistics}, pages 208--214.

\bibitem[Crump et~al., 2009]{crump2009dealing}
Crump, R.~K., Hotz, V.~J., Imbens, G.~W., and Mitnik, O.~A. (2009).
\newblock Dealing with limited overlap in estimation of average treatment
  effects.
\newblock {\em Biometrika}, 96(1):187--199.

\bibitem[Dimakopoulou et~al., 2017]{dimakopoulou2017estimation}
Dimakopoulou, M., Zhou, Z., Athey, S., and Imbens, G. (2017).
\newblock Estimation considerations in contextual bandits.
\newblock {\em arXiv preprint arXiv:1711.07077}.

\bibitem[Dimakopoulou et~al., 2019]{dimakopoulou2019balanced}
Dimakopoulou, M., Zhou, Z., Athey, S., and Imbens, G. (2019).
\newblock Balanced linear contextual bandits.
\newblock In {\em Proceedings of the AAAI Conference on Artificial
  Intelligence}, volume~33, pages 3445--3453.

\bibitem[Filippi et~al., 2010]{FCGS2010}
Filippi, S., Cappe, O., Garivier, A., and Szepesv{\'a}ri, C. (2010).
\newblock Parametric bandits: The generalized linear case.
\newblock In {\em Advances in Neural Information Processing Systems}, pages
  586--594.

\bibitem[Garivier and Capp{\'e}, 2011]{garivier2011kl}
Garivier, A. and Capp{\'e}, O. (2011).
\newblock The kl-ucb algorithm for bounded stochastic bandits and beyond.
\newblock In {\em Proceedings of the 24th annual conference on learning
  theory}, pages 359--376. JMLR Workshop and Conference Proceedings.

\bibitem[Garivier and Moulines, 2011]{garivier2011upper}
Garivier, A. and Moulines, E. (2011).
\newblock On upper-confidence bound policies for switching bandit problems.
\newblock In {\em International Conference on Algorithmic Learning Theory},
  pages 174--188. Springer.

\bibitem[Ghavamzadeh et~al., 2015]{ghavamzadeh2015bayesian}
Ghavamzadeh, M., Mannor, S., Pineau, J., Tamar, A., et~al. (2015).
\newblock Bayesian reinforcement learning: A survey.
\newblock {\em Foundations and Trends{\textregistered} in Machine Learning},
  8(5-6):359--483.

\bibitem[Graepel et~al., 2010]{graepel2010web}
Graepel, T., Candela, J.~Q., Borchert, T., and Herbrich, R. (2010).
\newblock Web-scale bayesian click-through rate prediction for sponsored search
  advertising in microsoft's bing search engine.
\newblock In {\em ICML}.

\bibitem[Hadad et~al., 2019]{hadad2019confidence}
Hadad, V., Hirshberg, D.~A., Zhan, R., Wager, S., and Athey, S. (2019).
\newblock Confidence intervals for policy evaluation in adaptive experiments.
\newblock {\em arXiv preprint arXiv:1911.02768}.

\bibitem[Hall and Heyde, 2014]{hall2014martingale}
Hall, P. and Heyde, C.~C. (2014).
\newblock {\em Martingale limit theory and its application}.
\newblock Academic press.

\bibitem[Imbens and Rubin, 2015]{imbens2015causal}
Imbens, G.~W. and Rubin, D.~B. (2015).
\newblock {\em Causal inference in statistics, social, and biomedical
  sciences}.
\newblock Cambridge University Press.

\bibitem[Jun et~al., 2017]{JBNW2017}
Jun, K.-S., Bhargava, A., Nowak, R., and Willett, R. (2017).
\newblock Scalable generalized linear bandits: Online computation and hashing.
\newblock In {\em Advances in Neural Information Processing Systems}, pages
  99--109.

\bibitem[Kaufmann et~al., 2012]{kaufmann2012thompson}
Kaufmann, E., Korda, N., and Munos, R. (2012).
\newblock Thompson sampling: An asymptotically optimal finite-time analysis.
\newblock In {\em International conference on algorithmic learning theory},
  pages 199--213. Springer.

\bibitem[Lai and Robbins, 1985]{lai1985asymptotically}
Lai, T.~L. and Robbins, H. (1985).
\newblock Asymptotically efficient adaptive allocation rules.
\newblock {\em Advances in applied mathematics}, 6(1):4--22.

\bibitem[Lattimore and Szepesv{\'a}ri, 2020]{lattimore2020bandit}
Lattimore, T. and Szepesv{\'a}ri, C. (2020).
\newblock {\em Bandit algorithms}.
\newblock Cambridge University Press.

\bibitem[Li and Chapelle, 2012]{li2012open}
Li, L. and Chapelle, O. (2012).
\newblock Open problem: Regret bounds for thompson sampling.
\newblock In {\em Conference on Learning Theory}, pages 43--1. JMLR Workshop
  and Conference Proceedings.

\bibitem[Li et~al., 2010]{LCLS2010}
Li, L., Chu, W., Langford, J., and Schapire, R.~E. (2010).
\newblock A contextual-bandit approach to personalized news article
  recommendation.
\newblock In {\em Proceedings of the 19th international conference on World
  wide web}, pages 661--670. ACM.

\bibitem[Li et~al., 2017]{LLZ2017}
Li, L., Lu, Y., and Zhou, D. (2017).
\newblock Provably optimal algorithms for generalized linear contextual
  bandits.
\newblock In {\em Proceedings of the 34th International Conference on Machine
  Learning-Volume 70}, pages 2071--2080. JMLR. org.

\bibitem[Luedtke and Van Der~Laan, 2016]{luedtke2016statistical}
Luedtke, A.~R. and Van Der~Laan, M.~J. (2016).
\newblock Statistical inference for the mean outcome under a possibly
  non-unique optimal treatment strategy.
\newblock {\em Annals of statistics}, 44(2):713.

\bibitem[May and Leslie, 2011]{may2011simulation}
May, B.~C. and Leslie, D.~S. (2011).
\newblock Simulation studies in optimistic bayesian sampling in
  contextual-bandit problems.
\newblock {\em Statistics Group, Department of Mathematics, University of
  Bristol}, 11(02).

\bibitem[Neel and Roth, 2018]{neel2018mitigating}
Neel, S. and Roth, A. (2018).
\newblock Mitigating bias in adaptive data gathering via differential privacy.
\newblock In {\em International Conference on Machine Learning}, pages
  3720--3729. PMLR.

\bibitem[Nie et~al., 2018]{nie2018adaptively}
Nie, X., Tian, X., Taylor, J., and Zou, J. (2018).
\newblock Why adaptively collected data have negative bias and how to correct
  for it.
\newblock In {\em International Conference on Artificial Intelligence and
  Statistics}, pages 1261--1269. PMLR.

\bibitem[Robbins, 1952]{robbins1952some}
Robbins, H. (1952).
\newblock Some aspects of the sequential design of experiments.
\newblock {\em Bulletin of the American Mathematical Society}, 58(5):527--535.

\bibitem[Russo, 2016]{russo2016simple}
Russo, D. (2016).
\newblock Simple bayesian algorithms for best arm identification.
\newblock In {\em Conference on Learning Theory}, pages 1417--1418. PMLR.

\bibitem[Russo and Van~Roy, 2014]{RV2014}
Russo, D. and Van~Roy, B. (2014).
\newblock Learning to optimize via posterior sampling.
\newblock {\em Mathematics of Operations Research}, 39(4):1221--1243.

\bibitem[Russo and Van~Roy, 2016]{russo2016information}
Russo, D. and Van~Roy, B. (2016).
\newblock An information-theoretic analysis of thompson sampling.
\newblock {\em The Journal of Machine Learning Research}, 17(1):2442--2471.

\bibitem[Russo et~al., 2017]{russo2017tutorial}
Russo, D., Van~Roy, B., Kazerouni, A., Osband, I., and Wen, Z. (2017).
\newblock A tutorial on thompson sampling.
\newblock {\em arXiv preprint arXiv:1707.02038}.

\bibitem[Scott, 2010]{scott2010modern}
Scott, S.~L. (2010).
\newblock A modern bayesian look at the multi-armed bandit.
\newblock {\em Applied Stochastic Models in Business and Industry},
  26(6):639--658.

\bibitem[Shin et~al., 2019]{shin2019bias}
Shin, J., Ramdas, A., and Rinaldo, A. (2019).
\newblock On the bias, risk and consistency of sample means in multi-armed
  bandits.
\newblock {\em arXiv preprint arXiv:1902.00746}.

\bibitem[Sutton and Barto, 2018]{sutton2018reinforcement}
Sutton, R.~S. and Barto, A.~G. (2018).
\newblock {\em Reinforcement learning: An introduction}.
\newblock MIT press.

\bibitem[Thompson, 1933]{thompson1933likelihood}
Thompson, W.~R. (1933).
\newblock On the likelihood that one unknown probability exceeds another in
  view of the evidence of two samples.
\newblock {\em Biometrika}, 25(3/4):285--294.

\bibitem[Xu et~al., 2013]{xu2013estimation}
Xu, M., Qin, T., and Liu, T.-Y. (2013).
\newblock Estimation bias in multi-armed bandit algorithms for search
  advertising.
\newblock {\em Advances in Neural Information Processing Systems},
  26:2400--2408.

\bibitem[Zhang et~al., 2020]{zhang2020inference}
Zhang, K.~W., Janson, L., and Murphy, S.~A. (2020).
\newblock Inference for batched bandits.
\newblock {\em arXiv preprint arXiv:2002.03217}.

\end{thebibliography}
